\documentclass[11pt,a4paper]{article}
\usepackage[hyperref]{emnlp-ijcnlp-2019}
\usepackage{times}
\usepackage{latexsym}

\usepackage{url}

\aclfinalcopy 

\usepackage{graphicx}
\usepackage[caption=false]{subfig}
\usepackage{color}
\usepackage{microtype}\usepackage{amsmath,amssymb,amsfonts}
\usepackage{mathtools}
\usepackage{mathrsfs}
\usepackage{bm}
\usepackage{booktabs, multirow}
\usepackage{makecell}
\usepackage{stfloats}
\usepackage{colortbl}
\definecolor{mygray}{gray}{.9}
\usepackage{hyperref}

\title{Modeling Conversation Structure and Temporal Dynamics for Jointly Predicting Rumor Stance and Veracity}

\author{Penghui Wei, Nan Xu, Wenji Mao\\
  $^{\dag}$SKL-MCCS, Institute of Automation, Chinese Academy of Sciences (CASIA) \\
  $^{\ddag}$University of Chinese Academy of Sciences \\
  {\tt \{weipenghui2016,xunan2015,wenji.mao\}@ia.ac.cn}}

\date{}

\begin{document}
\maketitle

\begin{abstract} 
Automatically verifying rumorous information has become an important and challenging task in natural language processing and social media analytics. Previous studies reveal that people's stances towards rumorous messages can provide indicative clues for identifying the veracity of rumors, and thus determining the stances of public reactions is a crucial preceding step for rumor veracity prediction. In this paper, we propose a hierarchical multi-task learning framework for jointly predicting rumor stance and veracity on Twitter, which consists of two components. The bottom component of our framework classifies the stances of tweets in a conversation discussing a rumor via modeling the structural property based on a novel graph convolutional network. The top component predicts the rumor veracity by exploiting the temporal dynamics of stance evolution. Experimental results on two benchmark datasets show that our method outperforms previous methods in both rumor stance classification and veracity prediction. 
\end{abstract}

\section{Introduction}
\label{intro}
Social media websites have become the main platform for users to browse information and share opinions, facilitating news dissemination greatly. However, the characteristics of social media also accelerate the rapid spread and dissemination of unverified information, i.e., rumors~\cite{shu2017fake}. The definition of \textit{rumor} is ``\textit{items of information that are unverified at the time of posting}''~\citep{zubiaga2018detection}. 
Ubiquitous false rumors bring about harmful effects, which has seriously affected public and individual lives, and caused panic in society~\cite{zhou2018fake,sharma2019combating}. 
Because online content is massive and debunking rumors manually is time-consuming, there is a great need for automatic methods to identify false rumors~\cite{oshikawa2018survey}. 

\begin{figure}[t]
\centering
\centerline{
\includegraphics[width=\columnwidth]{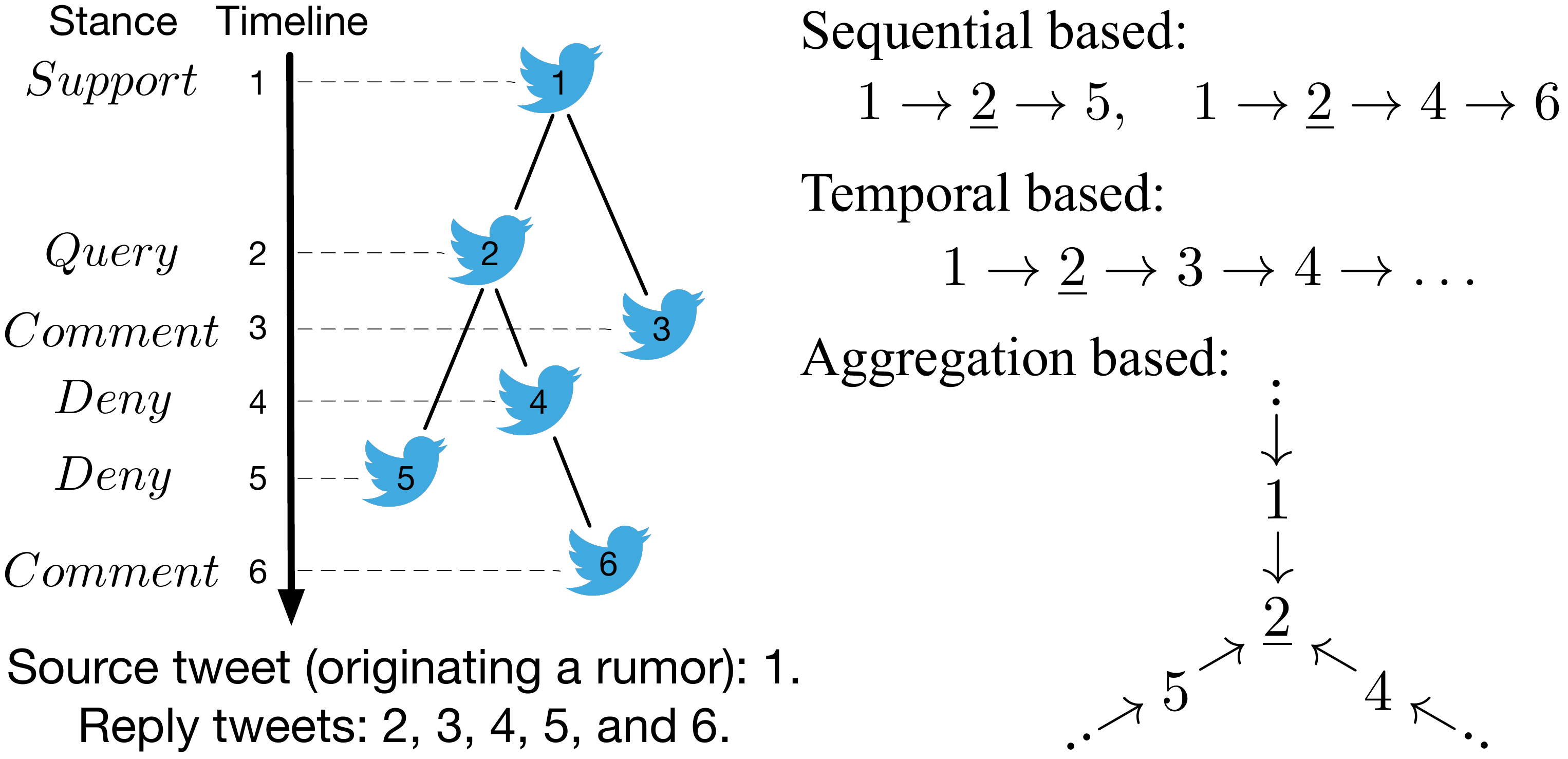}
}
\caption{A conversation thread discussing the rumorous tweet ``1''. Three different perspectives for learning the stance feature of the reply tweet ``2'' are illustrated.}
\label{example}
\end{figure}

Previous studies have observed that public stances towards rumorous messages are crucial signals to detect trending rumors~\cite{qazvinian2011rumor,zhao2015enquiring} and indicate the veracity of them~\cite{mendoza2010twitter,procter2013reading,liu2015real,jin2016news,glenski2018identifying}. 
Therefore, stance classification towards rumors is viewed as an important preceding step of rumor veracity prediction, especially in the context of Twitter conversations~\cite{zubiaga2016stance}. 

The state-of-the-art methods for rumor stance classification are proposed to model the sequential property~\cite{kochkina2017turing} or the temporal property~\cite{veyseh2017temporal} of a Twitter conversation thread. 
In this paper, we propose a new perspective based on structural property: learning tweet representations through aggregating information from their neighboring tweets. 
Intuitively, a tweet's nearer neighbors in its conversation thread are more informative than farther neighbors because the replying relationships of them are closer, and their stance expressions can help classify the stance of the center tweet (e.g., in Figure~\ref{example}, tweets ``1'', ``4'' and ``5'' are the one-hop neighbors of the tweet ``2'', and their influences on predicting the stance of ``2'' are larger than that of the two-hop neighbor ``3''). 
To achieve this, we represent both tweet contents and conversation structures into a latent space using a graph convolutional network (GCN)~\cite{kipf2016semi}, aiming to learn stance feature for each tweet by aggregating its neighbors' features. 
Compared with the sequential and temporal based methods, our aggregation based method leverages the intrinsic structural property in conversations to learn tweet representations. 

\begin{figure*}[t]
\centering
\centerline{
\includegraphics[width=1.8\columnwidth]{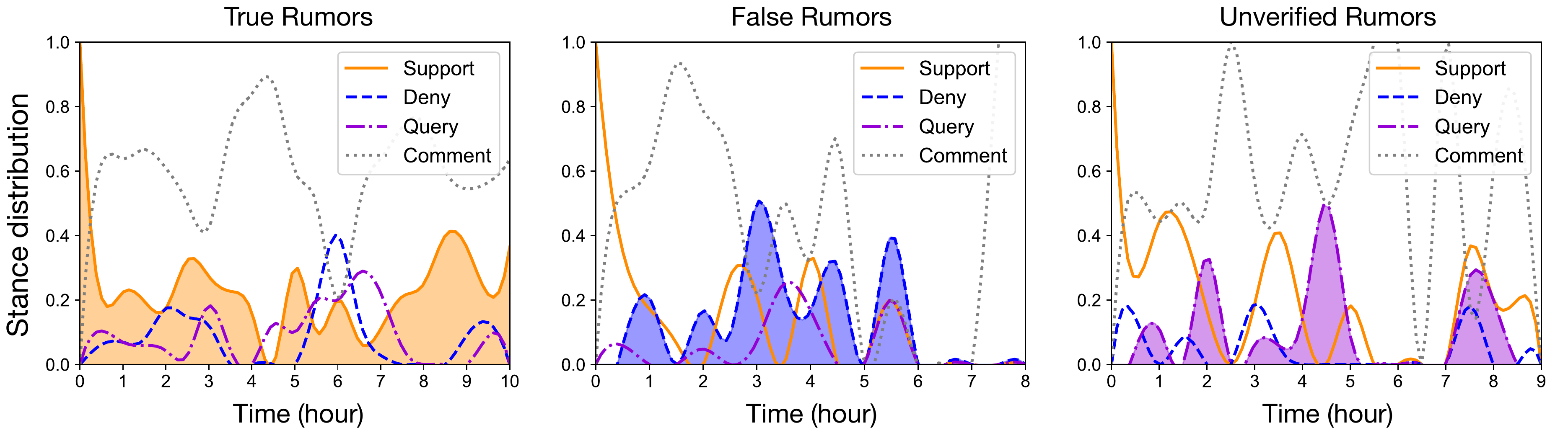}
}
\caption{Stance distributions of tweets discussing $true$ rumors, $false$ rumors, and $unverified$ rumors, respectively (Better viewed in color). The horizontal axis is the spreading time of the first rumor. It is visualized based on SemEval-2017 task 8 dataset~\cite{derczynski2017semeval}. All tweets are relevant to the event ``Ottawa Shooting''. }
\label{stance}
\end{figure*}

After determining the stances of people's reactions, another challenge is how we can utilize public stances to predict rumor veracity accurately. 
We observe that the temporal dynamics of public stances can indicate rumor veracity. 
Figure~\ref{stance} illustrates the stance distributions of tweets discussing $true$ rumors, $false$ rumors, and $unverified$ rumors, respectively. 
As we can see, $supporting$ stance dominates the inception phase of spreading. 
However, as time goes by, the proportion of $denying$ tweets towards $false$ rumors increases quite significantly. Meanwhile, the proportion of $querying$ tweets towards $unverified$ rumors also shows an upward trend. 
Based on this observation, we propose to model the temporal dynamics of stance evolution with a recurrent neural network (RNN), capturing the crucial signals containing in stance features for effective veracity prediction. 

Further, most existing methods tackle stance classification and veracity prediction separately, which is suboptimal and limits the generalization of models. As shown previously, they are two closely related tasks in which stance classification can provide indicative clues to facilitate veracity prediction. Thus, these two tasks can be jointly learned to make better use of their interrelation. 

Based on the above considerations, in this paper, we propose a hierarchical multi-task learning framework for jointly predicting rumor stance and veracity, which achieves deep integration between the preceding task (stance classification) and the subsequent task (veracity prediction).  
The bottom component of our framework classifies the stances of tweets in a conversation discussing a rumor via aggregation-based structure modeling, and we design a novel graph convolution operation customized for conversation structures. 
The top component predicts rumor veracity by exploiting the temporal dynamics of stance evolution, taking both content features and stance features learned by the bottom component into account. 
Two components are jointly trained to utilize the interrelation between the two tasks for learning more powerful feature representations. 

The contributions of this work are as follows. 

$\bullet$ We propose a hierarchical framework to tackle rumor stance classification and veracity prediction jointly, exploiting both structural characteristic and temporal dynamics in rumor spreading process. 

$\bullet$ We design a novel graph convolution operation customized to encode conversation structures for learning stance features. To our knowledge, we are the first to employ graph convolution for modeling the structural property of Twitter conversations. 

$\bullet$ Experimental results on two benchmark datasets verify that our hierarchical framework performs better than existing methods in both rumor stance classification and veracity prediction.

\section{Related Work}
\textbf{Rumor Stance Classification}\quad Stance analysis has been widely studied in online debate forums~\cite{Somasundaran2009Recognizing,hasan2013extra}, and recently has attracted increasing attention in different contexts~\cite{mohammad2016semeval,augenstein2016stance,ferreira2016emergent,mohtarami2018automatic}. 
After the pioneering studies on stance classification towards rumors in social media~\cite{mendoza2010twitter,qazvinian2011rumor,procter2013reading}, linguistic feature~\cite{hamidian2016rumor,zeng2016unconfirmed} and point process based methods~\cite{lukasik2016hawkes,lukasik2019gaussian} have been developed.

Recent work has focused on Twitter conversations discussing rumors.~\citet{zubiaga2016stance} proposed to capture the sequential property of conversations with linear-chain CRF, and also used a tree-structured CRF to consider the conversation structure as a whole.~\citet{aker2017simple} developed a novel feature set that scores the level of users' confidence.~\citet{pamungkas2018stance} designed affective and dialogue-act features to cover various facets of affect.~\citet{giasemidis2018semi} proposed a semi-supervised method that propagates the stance labels on similarity graph. Beyond feature-based methods,~\citet{kochkina2017turing} utilized an LSTM to model the sequential branches in a conversation, and their system ranked the first in SemEval-2017 task 8.~\citet{veyseh2017temporal} adopted attention to model the temporal property of a conversation and achieved the state-of-the-art performance. 

\textbf{Rumor Veracity Prediction}\quad Previous studies have proposed methods based on various features such as linguistics, time series and propagation structures~\cite{castillo2011information,kwon2013prominent,wu2015false,ma2017detect}. Neural networks show the effectiveness of modeling time series~\cite{ma2016detecting,ruchansky2017csi} and propagation paths~\cite{liu2018early}. 
~\citet{ma2018rumor}'s model adopted recursive neural networks to incorporate structure information into tweet representations and outperformed previous methods. 

Some studies utilized stance labels as the input feature of veracity classifiers to improve the performance~\cite{liu2015real,enayet2017niletmrg}.~\citet{dungs2018can} proposed to recognize the temporal patterns of true and false rumors' stances by two hidden Markov models (HMMs). Unlike their solution, our method learns discriminative features of stance evolution with an RNN. Moreover, our method jointly predicts stance and veracity by exploiting both structural and temporal characteristics, whereas HMMs need stance labels as the input sequence of observations.

\textbf{Joint Predictions of Rumor Stance and Veracity}\quad Several work has addressed the problem of jointly predicting rumor stance and veracity. These studies adopted multi-task learning to jointly train two tasks~\cite{ma2018detect,kochkina2018all,poddar2018predicting} and learned shared representations with parameter-sharing. 
Compared with such solutions based on ``parallel'' architectures, our method is deployed in a hierarchical fashion that encodes conversation structures to learn more powerful stance features by the bottom component, and models stance evolution by the top component, achieving deep integration between the two tasks' feature learning. 

\section{Problem Definition}
Consider a Twitter conversation thread $\mathcal C$ which consists of a source tweet $t_1$ (originating a rumor) and a number of reply tweets $\{t_2,t_3,\ldots,t_{|\mathcal C|}\}$ that respond $t_1$ directly or indirectly, and each tweet $t_i$ ($i\in [1, |\mathcal C|]$) expresses its stance towards the rumor. The thread $\mathcal C$ is a tree structure, in which the source tweet $t_1$ is the root node, and the replying relationships among tweets form the edges.\footnote{We consider each replying relationship between two tweets in a conversation as an undirected edge in the tree.} 

This paper focuses on two tasks. The first task is rumor stance classification, aiming to determine the stance of each tweet in $\mathcal C$, which belongs to $\{supporting,denying,querying,commenting\}$. The second task is rumor veracity prediction, with the aim of identifying the veracity of the rumor, belonging to $\{true,false,unverified\}$.

\begin{figure*}[t]
\centering
\centerline{
\includegraphics[width=2.0\columnwidth]{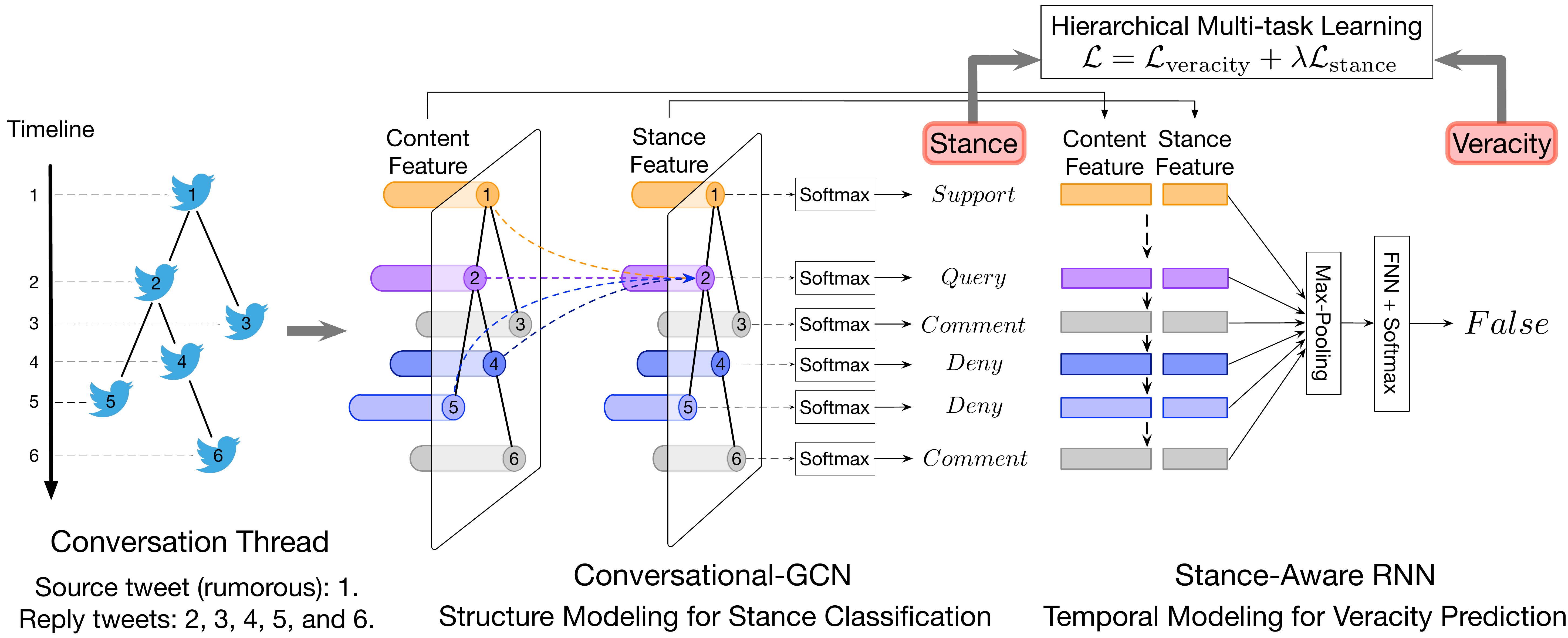}
}
\caption{Overall architecture of our proposed framework for joint predictions of rumor stance and veracity. In this illustration, the number of GCN layers is one. The information aggregation process for the tweet $t_2$ based on original graph convolution operation (Eq. (\ref{eq1})) is detailed.}
\label{overview}
\end{figure*}

\section{Proposed Method}
We propose a Hierarchical multi-task learning framework for jointly Predicting rumor Stance and Veracity (named \textit{Hierarchical-PSV}). Figure~\ref{overview} illustrates its overall architecture that is composed of two components. The bottom component is to classify the stances of tweets in a conversation thread, which learns stance features via encoding conversation structure using a customized graph convolutional network (named \textit{Conversational-GCN}). The top component is to predict the rumor's veracity, which takes the learned features from the bottom component into account and models the temporal dynamics of stance evolution with a recurrent neural network (named \textit{Stance-Aware RNN}). 

\subsection{Conversational-GCN: Aggregation-based Structure Modeling for Stance Prediction}
Now we detail \textit{Conversational-GCN}, the bottom component of our framework. 
We first adopt a bidirectional GRU (BGRU)~\cite{cho2014gru} layer to learn the content feature for each tweet in the thread $\mathcal C$. For a tweet $t_i$ ($i\in[1,|\mathcal C|]$), we run the BGRU over its word embedding sequence, and use the final step's hidden vector to represent the tweet. The content feature representation of $t_i$ is denoted as $\bm c_i\in\mathbb R^{d}$, where $d$ is the output size of the BGRU. 

As we mentioned in Section~\ref{intro}, the stance expressions of a tweet $t_i$'s nearer neighbors can provide more informative signals than farther neighbors for learning $t_i$'s stance feature. Based on the above intuition, we model the structural property of the conversation thread $\mathcal C$ to learn stance feature representation for each tweet in $\mathcal C$. 
To this end, we encode structural contexts to improve tweet representations by aggregating information from neighboring tweets with a graph convolutional network (GCN)~\cite{kipf2016semi}. 

Formally, the conversation $\mathcal C$'s structure can be represented by a graph $\mathcal C_{G}=\langle \mathcal T, \mathcal E \rangle$, where $\mathcal T=\{t_i\}_{i=1}^{|\mathcal C|}$ denotes the node set (i.e., tweets in the conversation), and $\mathcal E$ denotes the edge set composed of all replying relationships among the tweets. We transform the edge set $\mathcal E$ to an adjacency matrix $\mathbf A\in\mathbb R^{|\mathcal C|\times |\mathcal C|}$, where $\mathbf A_{ij}=\mathbf A_{ji}=1$ if the tweet $t_i$ directly replies the tweet $t_j$ or $i=j$. In one GCN layer, the graph convolution operation for one tweet $t_i$ on $\mathcal C_G$ is defined as:
\begin{equation}
\bm h_i^{\text{out}} = \tanh \left(\sum_{j\in\{j\mid \hat{\mathbf A}_{ij}\neq 0\}}{\hat{\mathbf A}_{ij}}\bm W^{\top}\bm h_j^{\text{in}}+\bm b\right)\,,
\label{eq1}
\end{equation}
where $\bm h_i^{\text{in}}\in\mathbb R^{d_{\text{in}}}$ and $\bm h_i^{\text{out}}\in\mathbb R^{d_{\text{out}}}$ denote the input and output feature representations of the tweet $t_i$ respectively. The convolution filter $\bm W\in\mathbb R^{d_{\text{in}}\times d_{\text{out}}}$ and the bias $\bm b\in\mathbb R^{d_{\text{out}}}$ are shared over all tweets in a conversation. We apply symmetric normalized transformation $\hat{\mathbf A}={\mathbf D}^{-\frac12}\mathbf A{\mathbf D}^{-\frac12}$ to avoid the scale changing of feature representations, where ${\mathbf D}$ is the degree matrix of $\mathbf A$, and $\{j\mid \hat{\mathbf A}_{ij}\neq 0\}$ contains $t_i$'s one-hop neighbors and $t_i$ itself. 

In this original graph convolution operation, given a tweet $t_i$, the receptive field for $t_i$ contains its one-hop neighbors and $t_i$ itself, and the aggregation level of two tweets $t_i$ and $t_j$ is dependent on $\hat{\mathbf A}_{ij}$. In the context of encoding conversation structures, we observe that such operation can be further improved for two issues. First, a tree-structured conversation may be very deep, which means that the receptive field of a GCN layer is restricted in our case. Although we can stack multiple GCN layers to expand the receptive field, it is still difficult to handle conversations with deep structures and increases the number of parameters. Second, the normalized matrix $\hat{\mathbf A}$ partly weakens the importance of the tweet $t_i$ itself. To address these issues, we design a novel graph convolution operation which is customized to encode conversation structures. Formally, it is implemented by modifying the matrix $\hat{\mathbf A}$ in Eq. (\ref{eq1}): 
\begin{equation}
\hat{\mathbf A}\leftarrow\hat{\mathbf A}\hat{\mathbf A}+\mathbf I\,,
\label{eq2}
\end{equation}
where the multiplication operation expands the receptive field of a GCN layer, and adding an identity matrix elevates the importance of $t_i$ itself. 

After defining the above graph convolution operation, we adopt an $L$-layer GCN to model conversation structures. The $l^{\text{th}}$ GCN layer ($l\in [1, L]$) computed over the entire conversation structure can be written as an efficient matrix operation:
\begin{equation}
\bm H^{(l)}=\tanh (\hat{\mathbf A}\bm H^{(l-1)}\bm W^{(l)}+\bm b^{(l)})\,,
\end{equation}
where $\bm H^{(l-1)}\in\mathbb R^{|\mathcal C|\times d_{l-1}}$ and $\bm H^{(l)}\in\mathbb R^{|\mathcal C|\times d_l}$ denote the input and output features of all tweets in the conversation $\mathcal C$ respectively. 

Specifically, the first GCN layer takes the content features of all tweets as input, i.e., $\bm H^{(0)}=(\bm c_1,\bm c_2,\ldots,\bm c_{|\mathcal C|})^{\top}\in\mathbb R^{|\mathcal C|\times d}$. The output of the last GCN layer represents the stance features of all tweets in the conversation, i.e., $\bm H^{(L)}=(\bm s_1,\bm s_2,\ldots,\bm s_{|\mathcal C|})^{\top}\in\mathbb R^{|\mathcal C|\times 4}$, where $\bm s_i$ is the unnormalized stance distribution of the tweet $t_i$. 

For each tweet $t_i$ in the conversation $\mathcal C$, we apply softmax to obtain its predicted stance distribution:
\begin{equation}
\hat{\bm s}_i= {\rm{softmax}}(\bm s_i)\in\mathbb R^4\,,\quad i\in[1,|\mathcal C|]\,.
\end{equation}

The ground-truth labels of stance classification supervise the learning process of \textit{Conversational-GCN}. The loss function of $\mathcal C$ for stance classification is computed by cross-entropy criterion:
\begin{equation}
\mathcal L_{\rm{stance}} = \frac{1}{|\mathcal C|}\sum_{i=1}^{|\mathcal C|}\left (-s_i^{\top}\log \hat{\bm s}_i\right )\,,
\end{equation}
where $s_i$ is a one-hot vector that denotes the stance label of the tweet $t_i$. For batch-wise training, the objective function for a batch is the averaged cross-entropy loss of all tweets in these conversations. 

In previous studies, GCNs are used to encode dependency trees~\cite{marcheggiani2017encoding,zhang2018graph} and cross-document relations~\cite{yasunaga2017graph,de2018question} for downstream tasks. Our work is the first to leverage GCNs for encoding conversation structures.

\subsection{Stance-Aware RNN: Temporal Dynamics Modeling for Veracity Prediction}
The top component, \textit{Stance-Aware RNN}, aims to capture the temporal dynamics of stance evolution in a conversation discussing a rumor. It integrates both content features and stance features learned from the bottom \textit{Conversational-GCN} to facilitate the veracity prediction of the rumor. 

Specifically, given a conversation thread $\mathcal C=\{t_1,t_2,\ldots,t_{|\mathcal C|}\}$ (where the tweets $t_*$ are ordered chronologically), we combine the content feature and the stance feature for each tweet, and adopt a GRU layer to model the temporal evolution:
\begin{equation}
\bm v_i = {\rm{GRU}}([\bm c_i; \bm s_i], \bm v_{i-1})\,,\quad i\in[1,|\mathcal C|]\,,
\end{equation}
where $[\cdot ;\cdot]$ denotes vector concatenation, and $(\bm v_1,\bm v_2,\ldots,\bm v_{|\mathcal C|})$ is the output sequence that represents the temporal feature. We then transform the sequence to a vector $\bm v$ by a max-pooling function that  captures the global information of stance evolution, and feed it into a one-layer feed-forward neural network (FNN) with softmax normalization to produce the predicted veracity distribution $\hat{\bm v}$:
\begin{equation}
\begin{aligned}
\bm v &= {\rm{max}}\text{-}{\rm{pooling}}(\bm v_1,\bm v_2,\ldots,\bm v_{|\mathcal C|})\,,\\
\hat{\bm v} &= {\rm{softmax}}({\rm{FNN}}(\bm v))\,.
\end{aligned}
\label{pooling}
\end{equation}

The loss function of $\mathcal C$ for veracity prediction is also computed by cross-entropy criterion:
\begin{equation}
\mathcal L_{\rm{veracity}} = -v^{\top}\log \hat{\bm v}\,,
\end{equation}
where $v$ denotes the veracity label of $\mathcal C$.

\subsection{Jointly Learning Two Tasks}
To leverage the interrelation between the preceding task (stance classification) and the subsequent task (veracity prediction), we jointly train two components in our framework. Specifically, we add two tasks' loss functions to obtain a joint loss function $\mathcal L$ (with a trade-off parameter $\lambda$), and optimize $\mathcal L$ to train our framework:
\begin{equation}
\label{loss}
\mathcal L=\mathcal L_{\rm{veracity}}+\lambda\mathcal L_{\rm{stance}}\,.
\end{equation}

In our \textit{Hierarchical-PSV}, the bottom component \textit{Conversational-GCN} learns content and stance features, and the top component \textit{Stance-Aware RNN} takes the learned features as input to further exploit temporal evolution for predicting rumor veracity. Our multi-task framework achieves deep integration of the feature representation learning process for the two closely related tasks.

\section{Experiments}
In this section, we first evaluate the performance of \textit{Conversational-GCN} on rumor stance classification and evaluate \textit{Hierarchical-PSV} on veracity prediction (Section~\ref{experimentalresult}). We then give a detailed analysis of our proposed method (Section~\ref{furtheranalysis}). 

\subsection{Data \& Evaluation Metric}
To evaluate our proposed method, we conduct experiments on two benchmark datasets. 

The first is \textbf{SemEval-2017 task 8}~\cite{derczynski2017semeval} dataset. It includes 325 rumorous conversation threads, and has been split into training, development and test sets. These threads cover ten events, and two events of that only appear in the test set. This dataset is used to evaluate both stance classification and veracity prediction tasks. 

The second is \textbf{PHEME} dataset~\cite{zubiaga2016analysing}. It provides 2,402 conversations covering nine events. 
Following previous work, we conduct leave-one-event-out cross-validation: in each fold, one event's conversations are used for testing, and all the rest events are used for training. The evaluation metric on this dataset is computed after integrating the outputs of all nine folds. 
Note that only a subset of this dataset has stance labels, and all conversations in this subset are already contained in SemEval-2017 task 8 dataset. Thus, PHEME dataset is used to evaluate veracity prediction task. 

Table~\ref{statistics} shows the statistics of two datasets. Because of the class-imbalanced problem, 
we use macro-averaged $F_1$ as the evaluation metric for two tasks. We also report accuracy for reference. 

\subsection{Implementation Details}
In all experiments, the number of GCN layers is set to $L=2$. We list the implementation details in Appendix A. 

\begin{table*}[t]
\small
\centering
\begin{tabular}{p{3em}<{\centering}cp{2.2em}<{\centering}cccccccc}
\toprule 
\multirow{2}*{\textbf{Dataset}} & \multirow{2}*{\# \textbf{Thread}} & \multirow{2}*{\textbf{Depth}} & \multirow{2}*{\# \textbf{Tweet}} & \multicolumn{4}{c}{\textbf{Stance Labels}} & \multicolumn{3}{c}{\textbf{Veracity Labels}}\\
\cmidrule(lr){5-8}\cmidrule(lr){9-11}
 & &  & & {\# \textit{support}} & {\# \textit{deny}} & {\# \textit{query}} & {\# \textit{comment}} & {\# \textit{true}} & {\# \textit{false}} & {\# \textit{unverified}} \\
\midrule
SemEval & 325 & 3.5 & 5,568 & 1,004 & 415 & 464 & 3,685 & 145 & 74 & 106\\
PHEME & 2,402 & 2.8 & 105,354 & \multicolumn{4}{c}{--} & 1,067 & 638 & 697 \\
\bottomrule
\end{tabular}
\caption{Statistics of two datasets. The column ``Depth'' denotes the average depth of all conversation threads.}
\label{statistics}
\end{table*}

\begin{table*}[t]
\small
\centering
\begin{tabular}{lcccccc}
\toprule
\multirow{2}*{\textbf{Method}} & \multicolumn{6}{c}{\textbf{Evaluation Metric}}\\
& Macro-$F_1$ & $F_{\text S}$ & $F_{\text D}$ & $F_{\text Q}$ & $F_{\text C}$ & Acc. \\  
\midrule
Affective Feature + SVM~\cite{pamungkas2018stance} & 0.470 & \textbf{0.410} & 0.000 & 0.580 & \textbf{0.880}  & 0.795 \\
BranchLSTM~\cite{kochkina2017turing} & 0.434 & 0.403 & 0.000 & 0.462 & 0.873  & 0.784 \\
TemporalAttention~\cite{veyseh2017temporal} & 0.482 & -- & -- & -- & --  & \textbf{0.820}\\
\rowcolor{mygray} Conversational-GCN (Ours, $L=2$)  & \textbf{0.499} & 0.311 & \textbf{0.194} & \textbf{0.646} & 0.847 & 0.751 \\
\bottomrule
\end{tabular}
\caption{Results of rumor stance classification. $F_{\text S}$, $F_{\text D}$, $F_{\text Q}$ and $F_{\text C}$ denote the $F_1$ scores of $supporting$, $denying$, $querying$ and $commenting$ classes respectively. ``--'' indicates that the original paper does not report the metric. 
}
\label{stanceresults}
\end{table*}

\subsection{Experimental Results}
\label{experimentalresult}
\subsubsection{Results: Rumor Stance Classification}
\textbf{Baselines}\quad We compare our \textit{Conversational-GCN} with the following methods in the literature: 

$\bullet$ \textit{Affective Feature + SVM}~\cite{pamungkas2018stance}\quad extracts affective and dialogue-act features for individual tweets, and then trains an SVM for classifying stances. 

$\bullet$  \textit{BranchLSTM}~\cite{kochkina2017turing}\quad is the winner of SemEval-2017 shared task 8 subtask A. It adopts an LSTM to model the sequential branches in a conversation thread. Before feeding branches into the LSTM, some additional hand-crafted features are used to enrich the tweet representations. 

$\bullet$  \textit{TemporalAttention}~\cite{veyseh2017temporal}\quad is the state-of-the-art method. It uses a tweet's ``neighbors in the conversation timeline'' as the context, and utilizes attention to model such temporal sequence for learning the weight of each neighbor. Extra hand-crafted features are also used.

\textbf{Performance Comparison}\quad Table~\ref{stanceresults} shows the results of different methods for rumor stance classification. Clearly, the macro-averaged $F_1$ of Conversational-GCN is better than all baselines. 

Especially, our method shows the effectiveness of determining  $denying$ stance, while other methods can not give any correct prediction for $denying$ class (the $F_{\text D}$ scores of them are equal to zero). Further, Conversational-GCN also achieves higher $F_1$ score for $querying$ stance ($F_{\text Q}$). Identifying $denying$ and $querying$ stances correctly is crucial for veracity prediction because they play the role of indicators for $false$ and $unverified$ rumors respectively (see Figure~\ref{stance}). Meanwhile, the class-imbalanced problem of data makes this a challenge. Conversational-GCN effectively encodes structural context for each tweet via aggregating information from its neighbors, learning powerful stance features without feature engineering. It is also more computationally efficient than sequential and temporal based methods. The information aggregations for all tweets in a conversation are worked in parallel and thus the running time is not sensitive to conversation's depth.

\begin{table*}[t]
\small
\centering
\begin{tabular}{clcccc}
\toprule
\multirow{2}*{\textbf{Setting}} & \multirow{2}*{\textbf{Method}} & \multicolumn{2}{c}{\textbf{SemEval dataset}} & \multicolumn{2}{c}{\textbf{PHEME dataset}}\\
\cmidrule(lr){3-4}\cmidrule(lr){5-6}
 & & Macro-$F_1$ & Acc. & Macro-$F_1$ & Acc. \\  
\midrule
\multirow{2}*{Single-task} & {TD-RvNN}~\cite{ma2018rumor} & 0.509 & 0.536 & 0.264 & 0.341\\
 & \cellcolor{mygray}{Hierarchical GCN-RNN (Ours)} & \cellcolor{mygray}{0.540} & \cellcolor{mygray}0.536 & \cellcolor{mygray}0.317 & \cellcolor{mygray}0.356\\
\midrule
\multirow{3}*{Multi-task} & {BranchLSTM+NileTMRG}~\cite{kochkina2018all} & 0.539 & 0.570 & 0.297 & 0.360\\
& {MTL2 (Veracity+Stance)}~\cite{kochkina2018all} & 0.558 & 0.571  & 0.318 & 0.357\\
 & \cellcolor{mygray}Hierarchical-PSV (Ours, $\lambda=1$)  & \cellcolor{mygray}\textbf{0.588} & \cellcolor{mygray}\textbf{0.643} & \cellcolor{mygray}\textbf{0.333} & \cellcolor{mygray}\textbf{0.361} \\
\bottomrule
\end{tabular}
\caption{Results of veracity prediction. Single-task setting means that stance labels cannot be used to train models.}
\label{veracityresults}
\end{table*}

\subsubsection{Results: Rumor Veracity Prediction}
To evaluate our framework \textit{Hierarchical-PSV}, we consider two groups of baselines: single-task and multi-task baselines. 

\textbf{Single-task Baselines}\quad In single-task setting, stance labels are not available. Only veracity labels can be used to supervise the training process.

$\bullet$ \textit{TD-RvNN}~\cite{ma2018rumor}\quad models the top-down tree structure using a recursive neural network for veracity classification. 

$\bullet$ \textit{Hierarchical GCN-RNN}\quad is the single-task variant of our framework: we optimize $\mathcal L_{\rm{veracity}}$ (i.e., $\lambda=0$ in Eq. (\ref{loss})) during training. Thus, the bottom Conversational-GCN only has indirect supervision (veracity labels) to learn stance features.

\textbf{Multi-task Baselines}\quad In multi-task setting, both stance labels and veracity labels are available for training. 

$\bullet$ \textit{BranchLSTM+NileTMRG}~\cite{kochkina2018all}\quad is a pipeline method, combining the winner systems of two subtasks in SemEval-2017 shared task 8. It first trains a BranchLSTM for stance classification, and then uses the predicted stance labels as extra features to train an SVM for veracity prediction~\cite{enayet2017niletmrg}. 

$\bullet$ \textit{MTL2 (Veracity+Stance)}~\cite{kochkina2018all}\quad is a multi-task learning method that adopts BranchLSTM as the shared block across tasks. Then, each task has a task-specific output layer, and two tasks are jointly learned.\footnote{\citet{kochkina2018all} also proposed \textit{MTL3} that jointly trains three tasks (plus rumor detection~\cite{zubiaga2017exploiting}). In our framework, we do not utilize data and labels from rumor detection task, and thus we choose \textit{MTL2 (Veracity+Stance)} as the state-of-the-art method for comparison.}

\textbf{Performance Comparison}\quad Table~\ref{veracityresults} shows the comparisons of different methods. By comparing single-task methods, Hierarchical GCN-RNN performs better than TD-RvNN, which indicates that our hierarchical framework can effectively model conversation structures to learn high-quality tweet representations. The recursive operation in TD-RvNN is performed in a fixed direction and runs over all tweets, thus may not obtain enough useful information. Moreover, the training speed of Hierarchical GCN-RNN is significantly faster than TD-RvNN: in the condition of batch-wise optimization for training one step over a batch containing 32 conversations, our method takes only 0.18 seconds, while TD-RvNN takes 5.02 seconds. 

Comparisons among multi-task methods show that two joint methods outperform the pipeline method (BranchLSTM+NileTMRG), indicating that jointly learning two tasks can improve the generalization through leveraging the interrelation between them. 
Further, compared with MTL2 which uses a ``parallel'' architecture to make predictions for two tasks, our Hierarchical-PSV performs better than MTL2. The hierarchical architecture is more effective to tackle the joint predictions of rumor stance and veracity, because it not only possesses the advantage of parameter-sharing but also offers deep integration of the feature representation learning process for the two tasks. Compared with Hierarchical GCN-RNN that does not use the supervision from stance classification task, Hierarchical-PSV provides a performance boost, which demonstrates that our framework benefits from the joint learning scheme. 

\subsection{Further Analysis and Discussions}
\label{furtheranalysis}
We conduct additional experiments to further demonstrate the effectiveness of our model.

\subsubsection{Effect of Customized Graph Convolution}
To show the effect of our customized graph convolution operation (Eq. (\ref{eq2})) for modeling conversation structures, we further compare it with the original graph convolution (Eq. (\ref{eq1}), named Original-GCN) on stance classification task. 

\begin{figure}[t]
\centering
\centerline{
\includegraphics[width=0.9\columnwidth]{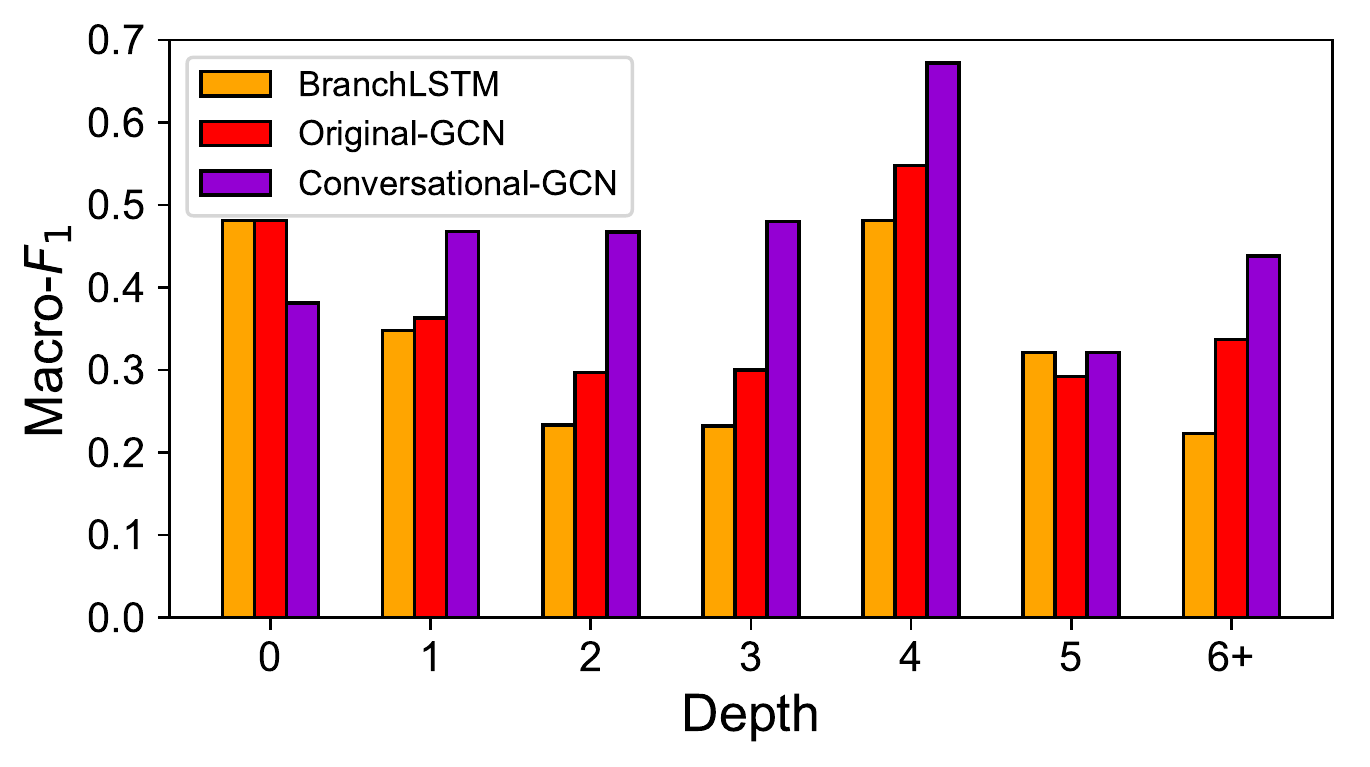}
}
\caption{Stance classification results w.r.t. different depths (see Appendix B for exact numerical numbers).}
\label{depthresults}
\end{figure}

Specifically, we cluster tweets in the test set according to their depths in the conversation threads (e.g., the cluster ``depth = 0'' consists of all source tweets in the test set). For BranchLSTM, Original-GCN and Conversational-GCN, we report their macro-averaged $F_1$ on each cluster in Figure~\ref{depthresults}. 

We observe that our Conversational-GCN outperforms Original-GCN and BranchLSTM significantly in most levels of depth. 
BranchLSTM may prefer to ``shallow'' tweets in a conversation because they often occur in multiple branches (e.g., in Figure~\ref{example}, the tweet ``2'' occurs in two branches and thus it will be modeled twice).
The results indicate that Conversational-GCN has advantage to identify stances of ``deep'' tweets in conversations.

\subsubsection{Ablation Tests}
\textbf{Effect of Stance Features}\quad 
To understand the importance of stance features for veracity prediction, we conduct an ablation study: we only input the content features of all tweets in a conversation to the top component RNN. It means that the RNN only models the temporal variation of tweet contents during spreading, but does not consider their stances and is not ``stance-aware''. Table~\ref{ablation} shows that ``-- stance features'' performs poorly, and thus the temporal modeling process benefits from the indicative signals provided by stance features. 
Hence, combining the low-level content features and the high-level stance features is crucial to improve rumor veracity prediction.

\begin{table}[t] 
\small
\centering
\begin{tabular}{lcc}
\toprule
\textbf{Method} &  Macro-$F_1$ & Acc.\\
\midrule
Hierarchical-PSV (full model) & \textbf{0.333} & \textbf{0.361}\\
\qquad -- stance features & 0.299 & 0.338\\
\cmidrule(lr){1-3}
\qquad -- GRU, + CNN  & 0.312 & 0.328\\
\qquad -- GRU  & 0.288 & 0.326\\
\bottomrule
\end{tabular}
\caption{Ablation tests of stance features and temporal modeling for veracity prediction on PHEME dataset.}
\label{ablation}
\end{table}

\begin{figure}[t]
\centering
\centerline{
\includegraphics[width=0.95\columnwidth]{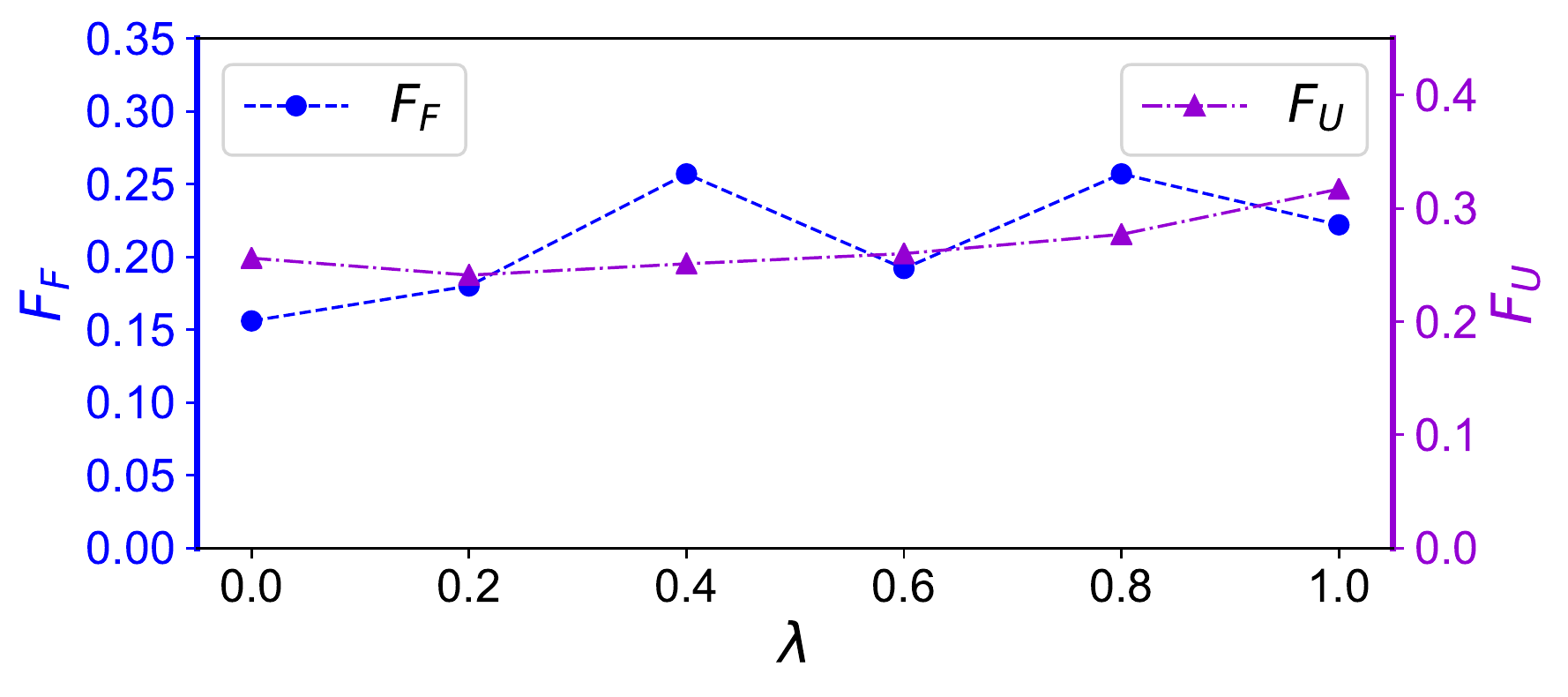}
}
\caption{Veracity prediction results v.s. various values of $\lambda$ on PHEME dataset. $F_{\text F}$ and $F_{\text U}$ denote the $F_1$ scores of $false$ and $unverified$ classes respectively.}
\label{interrelation}
\end{figure}

\textbf{Effect of Temporal Evolution Modeling}\quad 
We modify the Stance-Aware RNN by two ways: (i) we replace the GRU  layer by a CNN that only captures local temporal information; (ii) we remove the GRU layer. Results in Table~\ref{ablation} verify that replacing or removing the GRU block hurts the performance, and thus modeling the stance evolution of public reactions towards a rumorous message is indeed necessary for effective veracity prediction.

\subsubsection{Interrelation of Stance and Veracity}

We vary the value of $\lambda$ in the joint loss $\mathcal L$ and train models with various $\lambda$ to show the interrelation between stance and veracity in Figure~\ref{interrelation}. As $\lambda$ increases from 0.0 to 1.0, the performance of identifying $false$ and $unverified$ rumors generally gains. 
Therefore, when the supervision signal of stance classification becomes strong, the learned stance features can produce more accurate clues for predicting rumor veracity.

\begin{figure}[t]
\centering
\centerline{
\includegraphics[width=\columnwidth]{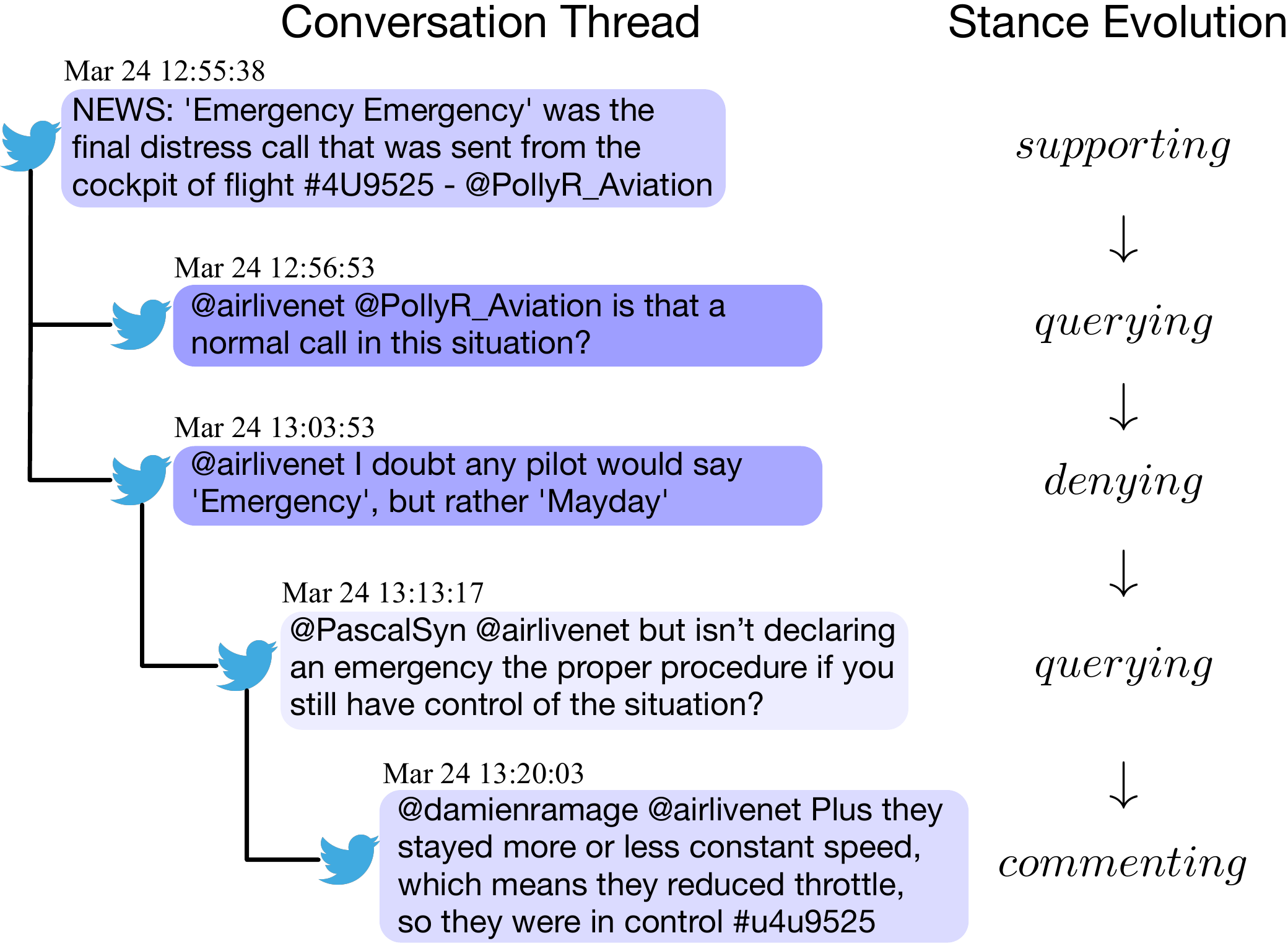}
}
\caption{Case study: a $false$ rumor. Each tweet is colored by the number of dimensions it contributes to $\bm v$ in the max-pooling operation (Eq. (\ref{pooling})). We show important tweets in the conversation and truncate others.}
\label{casestudy}
\end{figure}

\subsection{Case Study}
Figure~\ref{casestudy} illustrates a $false$ rumor identified by our model. We can observe that the stances of reply tweets present a typical temporal pattern ``$supporting\to querying\to denying$''. Our model captures such stance evolution with RNN and predicts its veracity correctly. Further, the visualization of tweets shows that the max-pooling operation catches informative tweets in the conversation. Hence, our framework can notice salience indicators of rumor veracity in the spreading process and combine them to give correct prediction.

\section{Conclusion}
We propose a hierarchical multi-task learning framework for jointly predicting rumor stance and veracity on Twitter. 
We design a new graph convolution operation, Conversational-GCN, to encode conversation structures for classifying stance, and then the top Stance-Aware RNN combines the learned features to model the temporal dynamics of stance evolution for veracity prediction. 
Experimental results verify that Conversational-GCN can handle deep conversation structures effectively, and our hierarchical framework performs much better than existing methods. 
In future work, we shall explore to incorporate external context~\cite{derczynski2017semeval,popat2018declare}, and extend our model to multi-lingual scenarios~\cite{wen2018cross}. Moreover, we shall investigate the diffusion process of rumors from social science perspective~\cite{vosoughi2018spread}, draw deeper insights from there and try to incorporate them into the model design.

\section*{Acknowledgments}
This work was supported in part by the National Key R\&D Program of China under Grant \#2016QY02D0305, NSFC Grants \#71621002, \#71472175, \#71974187 and \#71602184, and Ministry of Health of China under Grant \#2017ZX10303401-002. We thank all the anonymous reviewers for their valuable comments. We also thank Qianqian Dong for her kind assistance.

\bibliography{rumorstance19}
\bibliographystyle{acl_natbib}

\appendix
\section{Implementation Details}
\subsection{Hyperparameters}
\textbf{SemEval-2017 task 8 dataset}\quad We pretrain 200-dimensional word embeddings by Skip-gram with negative sampling~\cite{mikolov2013distributed}, and they are fixed during the training process. We set the dimension of content feature to 200. We use a two-layer GCN, and the output sizes of two layers are 200 and 4, respectively. 
The max-pooling function in Eq. (7) is to select the maximum value of each dimension, and thus the size of $\bm v$ is equal to that of $\bm v_i$. The trade-off parameter $\lambda$ is set to 1. 
We add dropout with 0.5 ratio to all but the last GCN layers and the FNN layer (used for veracity prediction). 
We train our model with 0.001 learning rate and 32 batch size using Adam~\cite{kingma2014adam}. For this dataset, after tuning hyperparameters on the development set, we merge the training and the development sets and re-train our model on the merged set.

\noindent\textbf{PHEME dataset}\quad We set the learning rate to 0.005 for accelerating the training process. Other configurations are same to that of SemEval dataset. Because only a subset of PHEME dataset contains stance labels, if a training conversation does not have stance labels, we will not compute its loss function of rumor stance classification task during the training process. 

\subsection{The CNN Layer in Ablation Study}
In Section 5.4.2, to demonstrate the effectiveness of modeling the temporal dynamics of stance evolution, we replace the GRU layer by a CNN layer that only captures local temporal information. 
Specifically, this CNN layer consists of a 1D convolution layer and a max-pooling function~\cite{kim2014convolutional}. We use three different filter windows: 2, 3 and 4. Each filter window has 100 feature maps. The output vector of this CNN layer is then fed into an FNN layer with softmax function to obtain the predicted veracity distribution.

\section{Numerical Results of Figure 4}
Table 5 shows the exact numerical numbers of the results in Figure 4.

\begin{table}[t]
\small
\centering
\begin{tabular}{cccc}
\toprule
\multirow{2}*{\textbf{Depth}} & \multicolumn{3}{c}{\textbf{Method}}\\
 & BranchLSTM & Original-GCN & Ours \\  
\midrule
0 & \textbf{0.481} & \textbf{0.481} & 0.381\\
1 & 0.348 & 0.363 & \textbf{0.468}\\
2 & 0.233 & 0.297 & \textbf{0.467}\\
3 & 0.232 & 0.300 & \textbf{0.480}\\
4 & 0.481 & 0.548 & \textbf{0.672}\\
5 & \textbf{0.321} & 0.292 & \textbf{0.321}\\
6+ & 0.223 & 0.337 & \textbf{0.438}\\
\bottomrule
\end{tabular}
\caption*{Table 5: Stance classification results w.r.t. different depths.}
\end{table}

\end{document}